\documentclass{article}

\usepackage{arxiv}
\usepackage[T1]{fontenc}

\usepackage{graphicx}
\usepackage{float}
\usepackage{cite}
\usepackage{todonotes}
\usepackage{comment}
\usepackage{flushend}
\usepackage{amsmath}
\usepackage{url}
\newcommand{\etal}{\textit{et al}. }

\begin{document}
\title{Pruning Convolutional Filters via Reinforcement Learning with Entropy Minimization}
%
%

\author{
 Mu\c sat Bogdan\\
  Transilvania University of Bra\c sov, Romania, \\ 
  Department of Electrical Engineering and Computer Science, \\
  \texttt{bogdan\_musat\_adrian@yahoo.com} \\
   \And
 Andonie R\u azvan \\
  Central Washington University, USA, \\ Department of Computer Science \\
  \texttt{razvan.andonie@cwu.edu} \\
}

\maketitle              

\begin{abstract}
Structural pruning has become an integral part of neural network optimization, used to achieve architectural configurations which can be deployed and run more efficiently on embedded devices. Previous results showed that pruning is possible with minimum performance loss by utilizing a reinforcement learning agent which makes decisions about the sparsity level of each neural layer by maximizing as a reward the accuracy of the network. We introduce a novel information-theoretic reward function which minimizes the spatial entropy of convolutional activations. This minimization ultimately acts as a proxy for maintaining accuracy, although these two criteria are not related in any way. Our method shows that there is another possibility to preserve accuracy without the need to directly optimize it in the agent's reward function. In our experiments, we were able to reduce the total number of FLOPS of multiple popular neural network architectures by 5-10$\times$, incurring minimal or no performance drop and being on par with the solution found by maximizing the accuracy.

\end{abstract}

\section{Introduction}
Modern convolutional neural networks (CNNs) emerged with the publication of AlexNet \cite{10.5555/2999134.2999257} in 2012, which paved the way for other architectures like VGG \cite{DBLP:journals/corr/SimonyanZ14a}, ResNet \cite{7780459} and EfficientNet \cite{DBLP:conf/icml/TanL19}. Although these networks posses a very high capacity and perform at a super human level, they are often overparametrized \cite{Oymak2020TowardMO}, which induces high latency and power consumption on battery powered devices. Techniques like pruning and quantization \cite{blalock2020state, gale2019state, Gholami2022ASO, 9578050, 9043731} have recently become very popular, since they can  generate power-efficient sub-versions of these overparametrized networks. 

While pruning deals with removing unimportant weights from a network by applying a certain heuristic, quantization operates by using less bits for weights and activations, thus speeding up overall computations. In our work we only focus on structured pruning, which translates into removing entire filters from a convolutional kernel. For this, we use an automated machine learning (AutoML) framework to select the most suited percentage of structured sparsity for each neural layer \cite{he2018amc}. 

AutoML is a powerful strategy used for many tasks like neural architecture search (NAS), hyperparameter search, data preparation, feature engineering \cite{HE2021106622, DBLP:journals/corr/abs-1810-13306}. The principle behind it is to automate manual searching tasks and find optimal solutions faster than we can manually do. Recently, AutoML was applied for network compression via pruning \cite{he2018amc}. By usage of a reinforcement learning (RL) agent \cite{journals/corr/LillicrapHPHETS15}, the system can automatically choose the sparsity percentage per layer, and then a magnitude-based pruning heuristic is applied, which removes the top percentage filters with the smallest magnitude. The reward criterion that the agent uses is the accuracy of the network obtained on a randomly chosen subset from the training set or the validation set at the end of the pruning phase.

We discovered that the accuracy of the network is not the only reward criterion that can be used for AutoML network compression. Our central contribution is an  information-theoretical reward function (entropy minimization) for the agent, which is completely different than the accuracy used in \cite{journals/corr/LillicrapHPHETS15}. We utilize this information-theoretical criterion for network pruning.

Generally, a low entropic value of a system denotes more certainty, while a high entropic value translates into more disorder. In neural networks, the cross-entropy between two probability distributions is very often used as a measure of error between the predicted output of a network and the true class distribution of the target. In contrast, the entropy of neural activations the hidden neural layers is seldom computed or used. This observation offered us the motivation for the current work. As such, we propose to analyze the impact of the entropy of all convolutional layers on network pruning. The question is if such an impact exists and, if true, how to use it for neural pruning. Since an ideal entropy value for the final output of the network is $0$ (i.e., to predict the correct class with 100\% confidence), our hypothesis is that the entropy of neural activations should be minimized in order to preserve essential information and also reduce uncertainty. 

The intriguing result of our work is the discovery of an interesting connection between entropy minimization and structural pruning. This could be related to the structural entropy measure recently introduced in \cite{Almog2019}, where "structural entropy refers to the level of heterogeneity of nodes in the network, with the premise that nodes that share functionality or attributes are more connected than others".

In practice, we utilize the AutoML framework from \cite{he2018amc} to sparsify a neural network and propose as an optimization reward criterion the minimization of the spatial entropy (as defined in \cite{e22121365}) at each convolutional layer. Through our experiments, we empirically show that this minimization acts as a proxy for maintaining accuracy. The novelty of our work consists in discovering that there are other, more principled approaches, to neural network pruning than directly optimize the accuracy of the agent's reward function.

The rest of the paper is structured as follows. Section \ref{related} describes related work. Section \ref{background} introduces the basic notations of spatial entropy, used in this paper. Section \ref{method} details our information-theoretical pruning method. The experimental results are described in Section \ref{experiments} and Section \ref{conclusions} concludes with final remarks.

\section{Related Work} \label{related}

In this section we summarize relevant previous work related to neural network pruning, AutoML and RL for pruning. 

\bigskip \noindent
\textbf{Pruning:} The main objective of pruning is the reduction of the total number of floating point operations per second (FLOPS) and parameters by removing redundant weights from the network. 

In the context of hardware accelerators, where computational availability might be limited, we argue that the ratio of preserved FLOPS for the same optimized architecture is a better measure of compression than simply counting the number of parameters after pruning, because two networks with the same number of parameters might posses totally different FLOPS counts, and as such the one with less FLOPS for the same should be more desirable. Only when memory space is an issue, a network with less parameters may be preferable.

The first approaches for neural pruning emerged in the '90s with the classical methods of optimal brain damage \cite{NIPS1989_6c9882bb} and optimal brain surgeon \cite{298572}. Since then, the importance of pruning was observed in improving training and inference time, better generalization. With the emergence of large deep neural networks, pruning became even more relevant and desirable, since modern network architectures are overparametrized and there is a lot of space for optimization. As such, a large suite of methods for pruning have been proposed in the last years. Comprehensive surveys on neural network pruning can be found in \cite{blalock2020state, gale2019state}.

Out of these methods, we highlight a few of them. Frankle \etal proposed the lottery ticket hypothesis \cite{frankle2018the}, representing a method to search for subnetworks derived from a randomly initialized network which, if trained from scratch, can attain the same performance as the original network and be sparse at the same time. In \cite{liu2018rethinking} the authors argued that the traditional pipeline of pruning: training, pruning, fine-tuning can result in networks with suboptimal performance and what is actually important is the architecture configuration itself resulted from pruning and not the weights that are maintained. As such, they showed that training a pruned network configuration from scratch can result in better accuracy than by fine-tuning it. The fine-tuning procedure is also studied in \cite{Renda2020Comparing} and they proposed two tricks to improve it. The first one is called weight rewinding which turns back weights which were not pruned to their earlier in time training values and continues retraining them using the original training schedule. The second trick is called learning rate rewinding which trains the final values of the unpruned weights using the same learning rate schedule as weight rewinding. Both tricks were shown to outperform classical fine-tuning. In \cite{Molchanov_2019_CVPR} the authors proposed a method to estimate the contribution of a filter to the final loss and iteratively remove those with smaller scores. To estimate a filter's importance they used two methods based on the first and second order Taylor expansions. \cite{pmlr-v119-evci20a} introduced "The Rigged Lottery" or RigL which trains sparse networks without the need of a "lucky" initialization. Their method updates the topology of a sparse network during training using parameter magnitudes and infrequent gradient calculations. They obtained better solutions than most dense-to-sparse training algorithms.

\bigskip \noindent
\textbf{AutoML:} AutoML has been widely adopted by the community to solve tasks where a huge amount of manual search would be involved. A comprehensive survey on AutoML can be found in \cite{HE2021106622, DBLP:journals/corr/abs-1810-13306}. We aim to present in the following some of the most important problems associated with AutoML. 

Data augmentation has become such an important part of an ML pipeline, that manually searching for strategies is no longer feasible or desirable. As such, methods like AutoAugment \cite{Cubuk_2019_CVPR}, RandAugment \cite{NEURIPS2020_d85b63ef} and Population Based Augmentation \cite{pmlr-v97-ho19b} have recently emerged in the literature and have been integrated in many training pipelines \cite{NEURIPS2020_d89a66c7, NEURIPS2020_f3ada80d, Xie_2020_CVPR}. Such methods are popular because, depending on the input images, they can automatically select between using multiple pre-existing augmentation techniques or choose the best hyperparameters for them.

Another important task is to search for optimal architectures  \cite{DBLP:journals/corr/ZophL16, pmlr-v80-pham18a, DBLP:journals/corr/abs-1806-09055, wrs-florea, WRSFloreaCNN}. Relevant architectures like EfficientNet B0 \cite{DBLP:conf/icml/TanL19} and AmoebaNet \cite{47354} were found by such techniques. Given the huge search space of neural layers' combinations, human inventiveness and intuition have fallen behind these methods which can exhaustively look up for better  solutions. Optimized networks were obtained this way, under specific hardware constraints \cite{DBLP:conf/iclr/CaiZH19}.

AutoML solutions for network pruning were also proposed \cite{he2018amc, 8354187}. These methods employ an RL agent to decide which individual weights or filters to drop, or the optimal amount of sparsity for each layer. It is known that some layers need more capacity \cite{blalock2020state}. Therefore, the uniform distribution approach for sparsity doesn't perform very well, and the sparsity has to be adapted by taking into consideration each layer's constraints. 

\bigskip \noindent
\textbf{Reinforcement learning for pruning:} An important result, the one on which our paper is based on, is by He \etal \cite{he2018amc}. They used a DDPG agent \cite{journals/corr/LillicrapHPHETS15} to choose the percentage of sparsity for each learnable layer, be it either convolutional or fully connected, by maximizing the accuracy of the network after pruning. For pruning convolutional layers, the filters with the lowest total magnitude were marked for sparsification, while for fully connected layers the smallest weights were the ones being discarded. This form of sparsification is considered to be structured, because weights are being dropped respecting a certain structure (i.e., a whole filter), and not being randomly dropped within a filter. This form of structured sparsification is more efficient, as whole filters can be removed from computations, while for the other form of sparsification, called random, some computations within filters have to be performed and others not, complicating the final arithmetic logic. 

Another important work that conducts pruning by means of an RL agent is \cite{NIPS2017_a51fb975}. The important difference compared to the previous work is that pruning is performed during runtime and adapted for the inputs. The assumption here is that some samples need more computational resources to be dealt with and thus the agent decides dynamically how much sparsity the network needs during inference. The main disadvantage is that an additional recurrent neural network (RNN) agent needs to run in order to sample actions during inference, while the previous method finds a fixed sub-network from the original one, which is also used during inference.

\section{Background: Spatial Entropy} \label{background}
Our goal is to minimize the spatial entropy of neural convolutional activations using the framework of AMC \cite{he2018amc} for structural sparsity. We also want to study whether this minimization can substitute direct accuracy optimization. If this assumption holds true, it might mean that there is a connection between pruning and entropy minimization. In the end, pruning unimportant filters ideally removes useless information. 

In \cite{e22121365} the notion of spatial entropy of saliency maps in CNNs was used to study the layer-wise and time-wise evolution of them \cite{e22121365,  IJCCC}. To compute the spatial entropy, the authors used the aura matrix entropy, initially defined in \cite{Volden1995}. In \cite{e22121365} it was noticed that the spatial entropy value decreases when going along the depth of a neural network as the process called semiotic superization takes place. Then, in \cite{IJCCC} the authors studied the dynamics of the spatial entropy in time, how it evolves during training, and found a link with fitting and compression present in the information bottleneck framework \cite{DBLP:journals/corr/Shwartz-ZivT17}. 

To make this paper self-contained, we describe below the mathematical formulations for computing the spatial entropy used in this work. Let us define the joint probability of features at spatial locations $(i, j)$ and $(i + k,\:j + l)$ to take the value $g$, respectively $g'$ as:

\begin{equation} \label{formula:2}
    p_{gg'}(k, l) = P(X_{i,\:j} = g, X_{i + k,\:j + l} = g')
\end{equation}
where $g$ and $g'$ are quantized convolutional activation values. In the Experiments section we tested using multiple bin sizes for quantization to verify the robustness of the procedure, and noticed that the proposed minimization produces good solutions for a wide range of bin size values. For each pair $(k,\:l)$ we define the entropy:

\begin{equation}
    H(k,\:l) = - \sum_{g} \sum_{g'} p_{gg'} (k,\:l) \log p_{gg'} (k,\:l)
\end{equation}
where the summations are over the number of bin sizes. A standardized relative measure of bivariate entropy is \cite{Journel1993}:

\begin{equation} \label{eq:4}
    H_R (k,\:l) = \frac{H(k,\:l) - H(0)}{H(0)} \in [0,\:1]
\end{equation}

The maximum entropy $H_R (k,\:l)=1$ corresponds to the case of two independent variables. $H(0)$ is the univariate entropy, which assumes all features as being independent, and we have $H(k,\:l) \geq H(0)$. 

Based on the relative entropy for $(k,\:l)$, the Spatial Disorder Entropy (SDE) for an $m \times n$ image $\textbf{X}$ was defined in \cite{Journel1993} as:

\begin{equation} \label{eq:5}
    H_{SDE} (\textbf X) \approx \frac{1}{mn} \sum_{i = 1}^m \sum_{j = 1}^n \sum_{k = 1}^m \sum_{l = 1}^n H_R (i - k,\:j - l)
\end{equation}

Since the complexity of SDE computation is high, we decided to use a simplified version - the Aura Matrix Entropy (AME, see \cite{Volden1995}), which only considers the second order neighbors from the SDE computation:

\begin{align} \label{eq:6}
\begin{split}
    H_{AME} (\textbf X) \approx \frac{1}{4} \bigg( H_R (-1,\:0) + H_R (0,\:-1) + H_R (1,\:0) + H_R (0,\:1) \bigg)
\end{split}
\end{align}

 Putting it all together, starting from a feature map obtained after the application of a convolutional layer, we compute the probabilities $p_{gg'}$ in equation  (\ref{formula:2}), and finally the AME in equation (\ref{eq:6}), which results in the spatial entropy quantity for a channel. We compute the mean of the spatial entropy values obtained from all the channels to get the final estimate for a particular convolutional activation layer. When computing the mean, we exclude channels which contain only $0$ values, which would produce a spatial entropy of $0$.

\section{Our Method} \label{method}

\begin{figure}[h]
    \centering
    \includegraphics[width=0.7\linewidth]{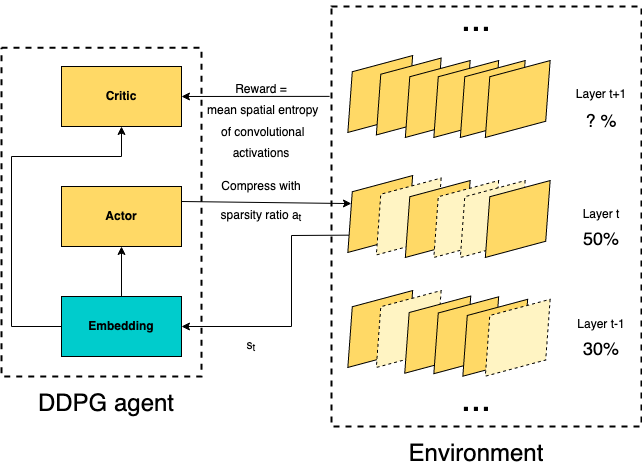}
    \caption{A DDPG agent is responsible with choosing the amount of sparsity $a_t$ applied to each layer, based on the state of that layer $s_t$, by considering multiple variables (see the original AMC paper for more details). After the agent decides the amount of sparsity for each layer, the environment computes a reward for the agent. In the original AMC formulation, this reward was the accuracy of the pruned network computed on a validation dataset. In our case, this reward consists in the mean of spatial entropies computed at multiple convolutional layer outputs. The agent's objective is to choose the amount of sparsity for each layer such that this mean entropy is minimized.}
    \label{fig:amc_entropy}
\end{figure}

This section describes our method, which is a modification of the AMC framework \cite{he2018amc} for structural pruning. The main difference between  our approach and \cite{he2018amc} is that our agent's reward function minimizes the spatial entropy of convolutional activations, instead of maximizing the accuracy of the model (see Figure \ref{fig:amc_entropy}).

The AMC framework is an AutoML tool for pruning which selects the percentage of sparsity for each layer of a neural network (layer by layer), then an algorithm based on $L_2$ magnitude marks the top percentage filters with lowest magnitude for removal. Since the accuracy of a model is a non-differentiable function for which gradients can not be computed and used during backpropagation, a RL technique has to be used to optimize this criterion, which will be treated as a reward function. Hence, the engine driving the percentage selection for pruning is a DDPG agent \cite{journals/corr/LillicrapHPHETS15}, trained via an actor critic technique \cite{NIPS1999_6449f44a}, using the accuracy computed on a separate dataset as a reward criterion. The reward function can be computed either on a split from the training dataset or the validation dataset. By pursuing optimal actions (with high reward), the agent will be rewarded correspondingly and encouraged to perform similar actions in the future, while generally discouraging actions with poor reward. 

AMC stores inputs and outputs for each layer at the very beginning of optimization by feed-forwarding a batch of input samples (called calibration samples). After pruning using the magnitude-based heuristic, the channels from the inputs of the calibration samples which have the same indexes as the discarded filters will be dropped as well. A least squares regression is applied to adjust the remaining weights to the new inputs and already stored outputs. After the AMC framework finds the best subnetwork configuration which maximizes the given reward function, and the weights have been adjusted as well via the least squares regression, the new network configuration is fine-tuned, as is standard in pruning literature. 

Different from the original AMC formulation, we modify the accuracy based reward by introducing a function which minimizes the average of the spatial entropies of convolutional activations. Our goal is to observe if entropy minimization can be used as a criterion in place of directly computing accuracy, establishing thus a potential interesting link between the fields of neural pruning and information theory. Because the framework of AMC tries to constantly increase the amount of reward it receives, and since the mean spatial entropy formulation we use is bounded between 0 and 1 \cite{Journel1993}, we subtract it from 1 in order to minimize the term. Thus, the optimization problem for the agent becomes finding the amount of sparsity for each layer which would eventually lead to minimize the spatial entropy. In order to compute the mean value (per layer) of the spatial entropy, we use the convolutional outputs from 100 samples, which of course represents only an estimate of the full dataset. Computing the mean spatial entropy using the full dataset would be computationally too costly, and as such resorting to a high smaller sample size is enough.

Our hypothesis is that by minimizing the spatial entropy, we can achieve on par or better results than when the goal is to maximize the accuracy. If this is the case, then we can establish an empirical connection between pruning and information theory, showing that by removing redundant information from a model we can achieve the same accuracy as when we directly try to maximize it. 

\section{Experiments} \label{experiments}

In this section we experimentally asses our hypothetical connection between information theory and pruning, verifying whether pruning achieved via AutoML, using minimization of spatial entropy for convolutional activations, can lead to a more compact model with similar accuracy. 

For training, we used the deep learning programming framework PyTorch \cite{NIPS2019_9015} (version 1.10.0) and the public implementation of AMC, modified to our needs.

We started by training a standard VGG-16 \cite{DBLP:journals/corr/SimonyanZ14a} on the CIFAR-10 dataset \cite{cifar10}. For that, we trained for 200 epochs using the SGD optimizer with a learning rate of $0.01$ and cosine annealing scheduler \cite{DBLP:conf/iclr/LoshchilovH17}. 

In order to establish a baseline to compare our method with, we used the original formulation of the AMC framework and optimize first the network using the accuracy criterion. To achieve a certain level of pruning, AMC pushes up the level of sparsity until only a predefined percentage of the total FLOPS are maintained. The ratio between the number of FLOPS after compression and the number of FLOPS before compression can measure indirectly the amount of sparsity in a network. Table \ref{table1} depicts the results for various FLOPS preservation percentages after fine-tuning on the CIFAR-10 dataset. For fine-tuning we used the same training optimizer and hyperparameters as described previously.

\begin{table}[h]
\centering
\caption{Accuracy for a VGG-16 network using the original AMC framework for different FLOPS preservation percentage}
\label{table1}
\begin{tabular}{ccccc} 
\hline
& Standard &  VGG-16 with  & VGG-16 with  & VGG-16 with \\
& VGG-16  & 50\% FLOPS & 20\% FLOPS & 10\% FLOPS \\
\hline
Accuracy & \textbf{93.58\%} & \textbf{93.85\%} & \textbf{93.26\%} & \textbf{92.18\%} \\
No. parameters & \textbf{14728266} & \textbf{4768242} & \textbf{912186} & \textbf{483402} \\
\hline
\end{tabular}
\end{table}

Our intuition is that entropy minimization should be the key in removing redundant information while preserving accuracy. Although, this should be the case, we performed a sanity test to check whether the opposite is also true (i.e., entropy maximization). We first performed some experiments for FLOPS preservation of 50\% and noticed that both entropy minimization and maximization achieve comparable accuracy results (before fine-tuning) on the mini validation set used for AMC agent training (Table \ref{table2}). 

\begin{table}[h]
\centering
\caption{Accuracy on mini split during AMC agent training for entropy minimization/maximization and FLOPS preservation of 50\%}
\label{table2}
\begin{tabular}{cc} 
\hline
Minimization & Maximization \\
\hline
\textbf{100\%} & \textbf{99.88\%} \\
\hline
\end{tabular}
\end{table}

Our next experiment used random rewards $\in [0, 1]$ to check whether 50\% FLOPS preservation is not a very simple task for the agent. On the mini split dataset used for agent training, we obtained an accuracy of $86\%$, concluding that even a reward which makes no sense can result in a network with decent performance. Therefore, we decided to use a scenario with 10\% preservation ratio in order to better stress test the framework and obtain more relevant results. Table \ref{table3} presents the accuracies obtained on CIFAR-10 after fine-tuning two networks obtained via AMC with entropy minimization/maximization and FLOPS preservation of 10\%. We notice that with entropy minimization we achieved the same performance as when accuracy is used as a reward. The solution found by this method has $10\times$ less FLOPS and $\approx 38 \times$ less parameters than the original VGG-16 network. For entropy maximization the framework produces a solution which has indeed fewer parameters, but uses the same number of FLOPS as the method with entropy minimization. We can see though that the resulting network architecture has a much poorer accuracy performance.

\begin{table}[h]
\centering
\caption{Experiments with entropy minimization and maximization and FLOPS preservation of 10\%. The accuracies are computed on the CIFAR-10 test set after fine-tuning. }
\label{table3}
\begin{tabular}{ccc} 
\hline
& Minimization & Maximization \\
\hline
Accuracy & \textbf{92.36\%} & \textbf{83.23\%} \\
No. parameters & \textbf{386442} & \textbf{91290} \\
\hline
\end{tabular}
\end{table}

In the above experiments, we used a bin size of 256 for quantizing the convolutional activations before computing the spatial entropy. Table \ref{table4} shows results for entropy minimization when using various bin sizes. It can be observed that our method is robust against a wide range of values, but the fewer bins results in faster computation - as expected. 

\begin{table}[h]
\centering
\caption{Accuracy on CIFAR-10 test set after fine-tuning. FLOPS preservation of 10\%, different bin size configurations}
\label{table4}
\begin{tabular}{cccccc} 
\hline
& 32 bins & 64 bins & 128 bins & 256 bins & 512 bins \\
\hline
Accuracy & \textbf{failed} & \textbf{92.43\%} & \textbf{92.31\%} & \textbf{92.36\%} & \textbf{91.97\%}\\
No. parameters & \textbf{failed} & \textbf{511938} & \textbf{619962} & \textbf{386442} & \textbf{578922} \\
\hline
\end{tabular}
\end{table}

The authors of \cite{liu2018rethinking} discovered that training from scratch a subnetwork configuration found via pruning can produce better results than fine-tuning the preserved weights. We verified this hypothesis with the architectures found by entropy minimization/maximization and noticed that in neither case training from scratch can produce better results (Table \ref{table5}). In our case, performance is better preserved when the weights are fine-tuned, because the remaining weights are properly adjusted using calibration samples, as described in Section \ref{method}.

\begin{table}[H]
\centering
\caption{Accuracy on CIFAR-10 test set when training a subnetwork from scratch}
\label{table5}
\begin{tabular}{ccccc} 
\hline
Acc. maximization & Acc. maximization & Acc. maximization & Entropy minimization \\
50\% FLOPS & 20\% FLOPS & 10\% FLOPS & 10\% FLOPS \\
\hline
\textbf{92.92\%} & \textbf{92.18\%} & \textbf{90.99\%} & \textbf{91.06\%}\\
\hline
\end{tabular}
\end{table}

In order to test the generality of our method for various other architectures, we repeated the same experiments for other popular networks: MobileNetV2 \cite{mobilenetv2} and ResNet50 \cite{7780459}. The results are depicted in Table \ref{table6}. Our method is on par with the original AMC framework for various architectures and FLOPS preservation percentages. The only noticeable drop in performance is for ResNet50, which was previously observed to contain less redundancy \cite{9578050} and was the most difficult to compress, even when using accuracy as a criterion.

\begin{table}[H]
\centering
\caption{Accuracy on CIFAR-10 test set with other architectures and various FLOPS preservation percentages}
\label{table6}
\begin{tabular}{cccccc} 
\hline
Architecture & Original & Accuracy & Accuracy & Entropy & Entropy\\
& performance & 50\% FLOPS & 20\% FLOPS & 50\% FLOPS & 20\% FLOPS\\
\hline
MobileNetV2 & \textbf{94.58}\% & \textbf{94.62\%} & \textbf{93.59}\% & \textbf{94.31}\% & \textbf{93.75\%} \\
ResNet50 & \textbf{95.21\%} & \textbf{95.34\%} & \textbf{95.09\%} & \textbf{95.27\%} & \textbf{94.27\%} \\
\hline
\end{tabular}
\end{table}

In our last experiment, we compared our method with other popular pruning methods. We used the ShrinkBench framework proposed in \cite{blalock2020state} to prune the same VGG16 network that we used in our experiments. From this framework, we selected the two most popular methods used for pruning: Global Magnitude Weight, which places together weights from all the layers and selects the ones with the largest magnitude and Local Magnitude Weight, which selects the weights with the largest magnitudes on a per layer basis. Table \ref{table7} shows the results obtained using ShrinkBench. It can be noticed that the accuracies resulted when using this framework are similar to what we obtain using AMC combined with our entropic criterion. One disadvantage of this method is that the type of pruning is unstructured, having the risk that when the network would be deployed to a specialized hardware, it might not reduce computational costs as much as when structured pruning is employed.

\begin{table}[h]
\centering
\caption{Accuracy for VGG16 on CIFAR-10 test set using the ShrinkBench framework}
\label{table7}
\begin{tabular}{ccccc} 
\hline
Original & Global & Global & Local & Local \\
Accuracy & 50\% FLOPS & 10\% FLOPS & 50\% FLOPS & 10\% FLOPS \\
\hline
\textbf{93.58\%} & \textbf{93.78\%} & \textbf{92.83\%}  & \textbf{93.27\%} & \textbf{92.34\%}\\
\hline
\end{tabular}
\end{table}

Using an information-theoretical optimization criterion, which aims to minimize entropy, we achieved the same performance as when we optimize directly the accuracy of the model. We were able to reduce the total number of FLOPS of a VGG-16 architecture by $10\times$ and the number of parameters by $\approx 38\times$, while incurring minimal accuracy drop, with similar results for other popular architectures. 

\section{Conclusions} \label{conclusions}

A standard neural network's output should ideally have a close to zero entropic value in order to confidently predict a class - not necessarily the correct one. Usually, this behavior is achieved by minimization of the cross-entropy between the network's output and the one hot encoding of the true class. This task can be sometimes burdensome, because internal layers of the network are not forced in any way to minimize the final entropy of the output layer. 

In our experiments, we explicitly forced the spatial entropy of internal convolutional activations to decrease with the goal to achieve neural pruning. According to our results, using the spatial entropy as an optimization criterion in an AutoML pruning framework, we can achieve good performance for an object recognition task, without directly optimizing the final evaluation metric (accuracy in this case). Because of the overparametrization of a neural network, removing  unessential information via entropy minimization helps reduce the model to its relevant (essential) components.

We established an interesting connection between information theory and neural pruning. Our result creates the premises for future applications in neural network pruning. In the future, we aim to explore novel ways in which the entropy can be used for the optimization of neural architectures. One such idea would be to include the entropy measure as a heuristic for selecting the channels to be preserve, instead of using the well known L2 magnitude. Another direction we aim to pursue is whether there is a link between pruning by entropy and semiotic aggregation, as described in \cite{e22121365}.

%
%
%
%

\end{document}